\documentclass[10pt,twocolumn,letterpaper]{article}

\usepackage[pagenumbers]{cvpr} 

\usepackage{graphicx}
\usepackage{amsmath}
\usepackage{amssymb}
\usepackage{booktabs}
\usepackage{lipsum}
\usepackage[toc,page]{appendix}
\usepackage{listings}

\usepackage{fancyvrb}
\usepackage{listings}
\usepackage{color}

\definecolor{dkgreen}{rgb}{0,0.6,0}
\definecolor{gray}{rgb}{0.5,0.5,0.5}
\definecolor{mauve}{rgb}{0.58,0,0.82}

\usepackage[pagebackref,breaklinks,colorlinks]{hyperref}

\usepackage[capitalize]{cleveref}
\crefname{section}{Sec.}{Secs.}
\Crefname{section}{Section}{Sections}
\Crefname{table}{Table}{Tables}
\crefname{table}{Tab.}{Tabs.}

\def\cvprPaperID{5430} 
\def\confName{CVPR}
\def\confYear{2023}

\begin{document}

\title{Parsing Line Segments of Floor Plan Images Using Graph Neural Networks}

\author{
Mingxiang Chen, Cihui Pan \\
Realsee \\
Beijing, China 100085 \\
{\tt\small \{chenmingxiang002, pancihui001\}@realsee.com} \\
}

\maketitle

\begin{abstract}
In this paper, we present a GNN-based Line Segment Parser (GLSP), which uses a junction heatmap to predict line segments' endpoints, and graph neural networks to extract line segments and their categories. Different from previous floor plan recognition methods, which rely on semantic segmentation, our proposed method is able to output vectorized line segment and requires less post-processing steps to be put into practical use.
Our experiments show that the methods outperform state-of-the-art line segment detection models on multi-class line segment detection tasks with floor plan images. In the paper, we use our floor plan dataset named Large-scale Residential Floor Plan data (LRFP). The dataset contains a total of 271,035 floor plan images. The label corresponding to each picture contains the scale information, the categories and outlines of rooms, and the endpoint positions of line segments such as doors, windows, and walls. Our augmentation method makes the dataset adaptable to the drawing styles of as many countries and regions as possible.
\end{abstract}


\begin{figure}[htb]
  \centering
  \includegraphics[width=1.0\linewidth]{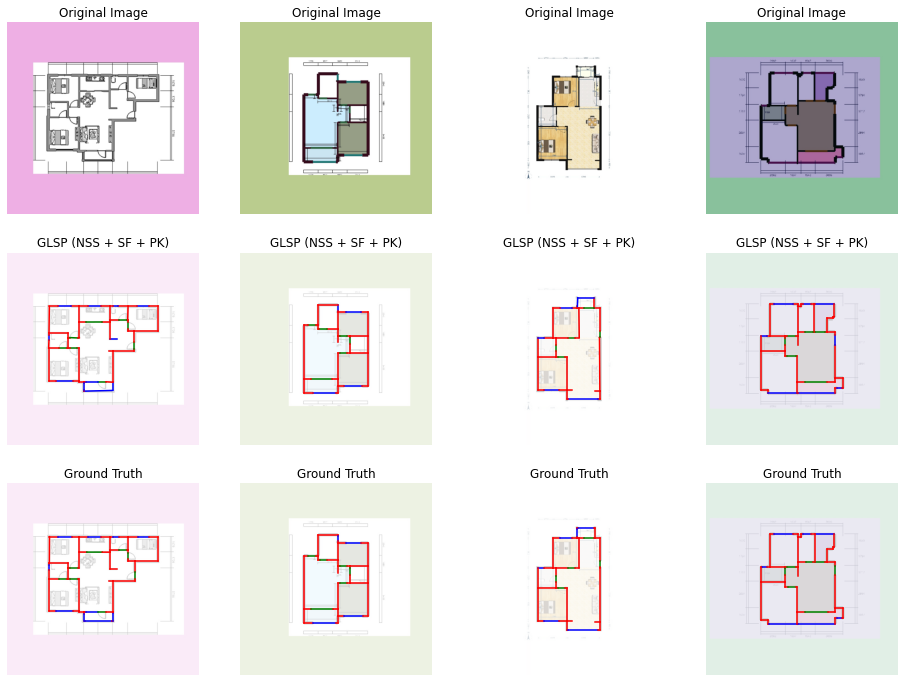}
  \caption{The proposed GLSP model can reliably translate images to a set of vectorized line segments with different types rather than semantic segmentation maps in previous studies \cite{Liu0KF17, DBLP:journals/corr/abs-1908-11025, Lv_2021_CVPR}. Augmentation methods are used so that the floor plans in the dataset have various styles. The walls, doors, and windows in the figures are represented by red, green, and blue line segments, respectively.}
  \label{fig:title_img}
\end{figure}

\section{Introduction}

Floor plan recognition has long been an active research field after deep learning has risen as a promising and stable method regarding computer vision problems. The task is straightforwardly designed, that is to recover the vector-graphic representation of a floor plan from a rasterized image, and re-enable further computing capabilities such as editing, synthesis, or analysis. In 2017, Liu \textit{et al.} \cite{Liu0KF17} proposed a method that combines deep learning with integer programming to identify key points, room outlines, and room categories. Some studies \cite{Surikov2020, Lv_2021_CVPR} show that using optical character recognition (OCR) or object detection methods for auxiliary judgment can further improve the accuracy. Although these methods show good results to some extent, they require heavy post-processing steps to be put into practical use.

In this paper, we propose a novel floor plan recognition method based on line segment detection (LSD) using Graph Neural Networks (GNN). While parsing the floor plans with two separate stages may introduce extra complexity, our method is able to extract vectorized line segments from floor plans rather than pixel-wise semantic segments. Despite the recent achievements made by deep learning in the field of LSD, two problems remain unsolved. First, line segments in floor plans have different categories such as doors, windows, or walls, while the detection methods proposed so far are not designed for multi-class line segments. Second, the algorithms performing well in natural scenes may not be the best choice in floor plan recognition tasks. For almost every blueprint images, including floor plans, the line segments are clearly and logically related to each other, which is different from the loose relationship between line segments in natural scenes.

Overall, our contributions are summarized as follows:

\begin{itemize}

\item We introduce the task of multi-class line segment detection into the field of floorplan recognition.

\item Our proposed method outputs vectorized results of structural elements and requires less post-processing steps to put the algorithm into practical use.

\item An attention-based graph neural network is used to capture the relationships between line segments accurately. The model achieves better performance compared to the state-of-the-art wireframe parsing model.


\end{itemize}

The paper is organized as follows. First, we introduce related works in Section \ref{sec: related works}. The details of our method are explained in Section \ref{sec: method}. The settings of experiments and their results are presented in Section \ref{sec: exp}. The conclusion is discussed in Section \ref{sec: conclusion}.

\section{Related Works} \label{sec: related works}

\textbf{Floor plan recognition and reconstruction} At present, many methods based on deep learning divide the problem of floor plan recognition and reconstruction into several more typical sub-problems, such as object detection, semantic segmentation, optical character recognition (OCR), etc. The system would integrate the recognition results of each model through a series of post-processing methods, and output standardized floor plans. For example, Liu \textit{et al.} \cite{Liu0KF17} use convolutional neural networks to identify the locations and types of junction points and use integer programming to output the information about walls and doors. The room types are recognized by a per-pixel semantic classification model. However, this method will not work if inclined walls are present in the floor plan since the types of each junction point are predefined. Zeng \textit{et al.} \cite{DBLP:journals/corr/abs-1908-11025} improves the accuracy of the semantic segmentation by using a room-boundary guided attention mechanism, while the corresponding post-processing methods are not proposed, so the results obtained are not vectorized floor plans. Surikov \textit{et al.} \cite{Surikov2020} use Faster R-CNN the object detection on floor plans. Lv \textit{et al.} \cite{Lv_2021_CVPR} improves the algorithm process based on \cite{Liu0KF17}, adding steps such as scale prediction, OCR, and object detection, which greatly improves the robustness and usability of the system.

\textbf{Datasets of floor plans} To the best of our knowledge, Raster-to-Vec \cite{Liu0KF17} is one of the earliest approaches trying to reconstruct floor plans from images. Its dataset contains 870 vector-graphics annotations. Rent3D \cite{ApartmentsCVPR15} is also a very popular open-source floor plan dataset containing 215 floor plan images. Recently, researchers have begun to use larger datasets for model training. Kalervo \textit{et al.} \cite{DBLP:journals/corr/abs-1904-01920} provides Cubi-Casa5K, including 5000 floor plan images from Finland. Lv \textit{et al.} \cite{Lv_2021_CVPR} mentioned Residential Floor Plan data (RFP) in their paper, which contains 7000 floor plans crawled from internet. However, the dataset is not open-source. Although the demand for larger-scale datasets is increasing without a doubt, it is difficult to obtain a large amount of floor plan data due to copyright or personal privacy protection. In addition, the existing floor plan datasets are generally only applicable to certain countries because of the drawing styles. Thus, even if the scale of some datasets such as RPLAN \cite{wu2019data} is sufficient to build large models, researchers from other places may be reluctant to use them.

\textbf{Line segment detection} Specifically, we use line segment detection methods as the pre-processing module in some of our baseline models. Edge detection \cite{canny1986computational,dollar2006supervised,dollar2013structured,martin2004learning,xie2015holistically} and perceptual grouping \cite{elder2002ecological,guil1995fast,smith1997susan} are classic methods often used by pioneers in this field. In addition, the method based on Hough transform \cite{duda1972use,furukawa2003accurate,guil1995fast,matas2000robust} is also a group of commonly used line segment detection methods based on traditional image processing. In the era of deep learning, the methods based on junction prediction represented by LCNN \cite{zhou2019end} and the methods based on dense prediction represented by AFM \cite{xue2019learning} have each shown excellent results and performance. HAWP \cite{xue2020holistically} combines the two on their basis, that is, to use the holistic attraction field map to propose a series of line segments, use junctions to fine-tune the results of the proposal or remove unreasonable line segments, and output the confidence of the refined line segments. Later, F-Clip \cite{dai2022fully} further optimizes the above model, abandons the two-stage paradigm, and improves both speed and accuracy. HAWPv3 \cite{xue2022hawp} explores the self-supervised learning paradigms based on HAWP and can be used as a good wireframe parser for the out-of-distribution images. Some researchers \cite{meng2020lgnn,zhang2019ppgnet} have proposed line segment detection methods based on graph networks. However, they do not perform better \cite{meng2020lgnn} than the above-mentioned two-phase parsing paradigm when detecting line segments in natural scenes. Recently, as the potential of the Transformer model being continuously explored, LETR \cite{xu2021line} builds a transformer-based end-to-end line segment detection algorithm by adding endpoint distance loss and coarse-to-fine decoding to the transformer model DETR \cite{carion2020end}, which is originally built for object detection.


\section{Method} \label{sec: method}

Similar to previous researches \cite{zhou2019end,xue2020holistically} on line segment detection, the floor plan representations are based on the notation of graph theory. A floor plan is defined on an undirected graph $\mathcal{G}=(\mathcal{P},\mathcal{A})$ where $i\in\mathcal{P}=\{1,2,...,n\}$ represents the $i$-th endpoint of all line segments, and $a_{i,j}\in\mathcal{A}$, where $i,j\in\mathcal{P},i\neq j$, represents the line segment from endpoint $i$ to $j$. For each endpoint $i$, the coordinate in the image space is represented by $p_i$. Different from line segment detection, the line segments in the floor plan have different categories (in this article are null, walls, doors, and windows), which are represented by $c_{i,j}$.

\begin{figure}[!t]
    \centering
    \includegraphics[width=.99\linewidth]{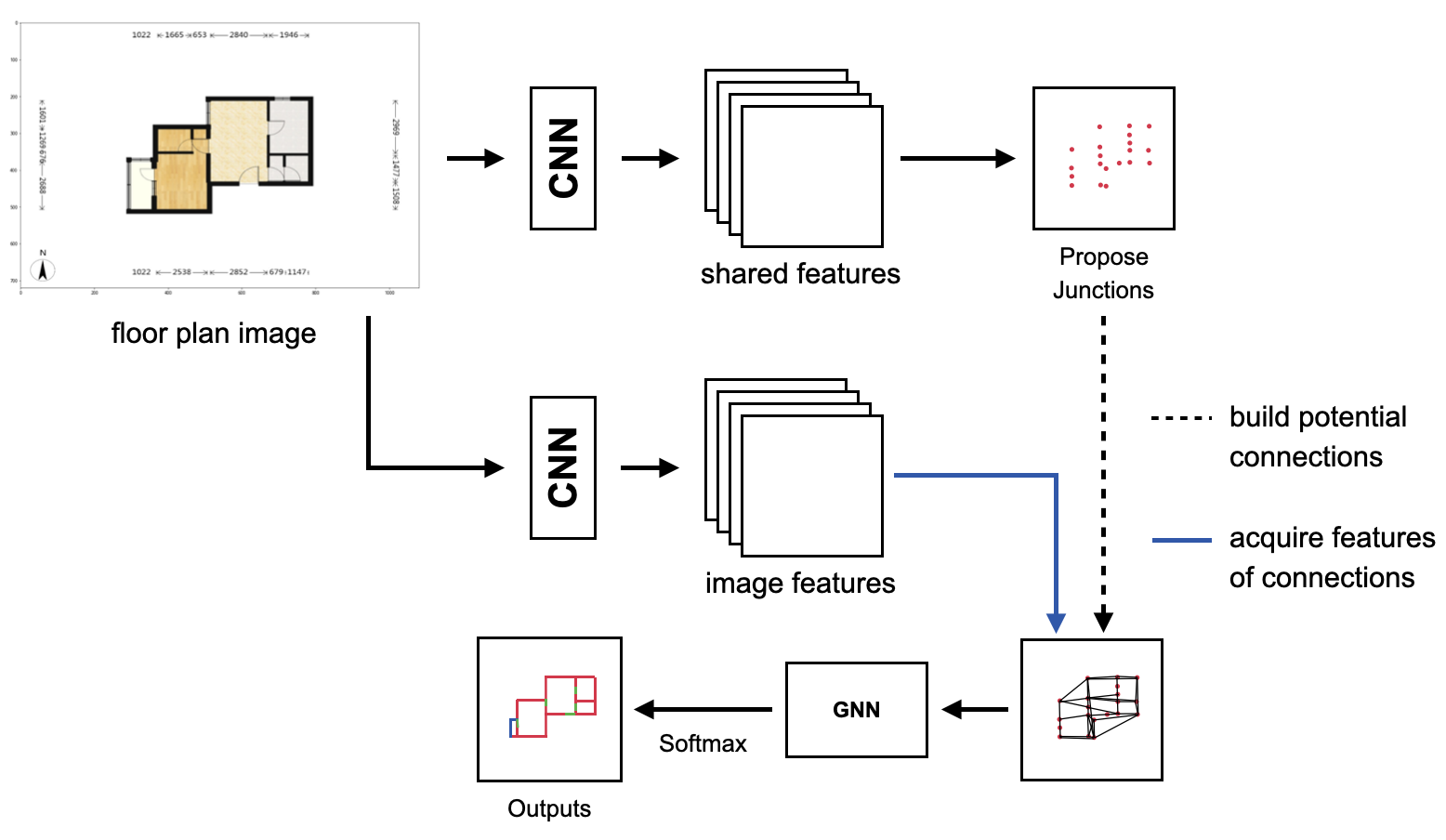}
    \caption{An overview of the model structure.}\label{Fig:model_structure_1}
\end{figure}

In this section we first introduce the dataset used for training and evaluation in Section \ref{subsec3: data}. Figure \ref{Fig:model_structure_1} illustrates an overview of our GNN-based Line Segment Parser (GLSP) architecture. For a floor plan image, we create two intermediate feature maps using identical backbone networks. One is used for junction detection (Section \ref{subsec3: junc}), and the another is used for building the features of potential connections in the graph (Section \ref{subsec331}). 
We use a graph neural network to classify the connections (Section \ref{subsec332} and \ref{subsec333}). Finally, the training strategies and the multi-task loss function are described in Section \ref{subsec3: loss}.

\subsection{Data Description} \label{subsec3: data}

All the floor plan images and labels in this dataset come from manual annotations by our 3D scanning operators. Each floor plan has been slightly modified, so users would not know their real locations. The houses corresponding to the floor plans are mostly Chinese urban residential buildings. The dataset is randomly split into training, validation, and test sets with 268,035, 1,500, and 1,500 floor plans, respectively. Each sample has 4 floor plan images including pictures with or without items of furniture and room labels. The images are saved in JPG format with a resolution of $1080 \times 720$. The annotations for floor plans include: 1) the scales of images represented by millimeters per pixel, 2) the information of lines including the starting and the ending points, thickness, and the category, choosing from \textit{wall}, \textit{door}, and \textit{window}, and 3) the information of rooms including the category and the contour. Please refer to the supplementary material for augmentation details and statistics about the dataset. 

\subsection{Junction Detection} \label{subsec3: junc}

We use the stacked Hourglass network \cite{newell2016stacked} as the feature extraction backbone for junction detection and the feature extraction module in the graph-building stage. The network is used in previous line segment detection researches \cite{zhou2019end,xue2020holistically,dai2022fully} and is also a commonly used model in human keypoint estimation. We choose the same settings as in \cite{xue2020holistically}, so that the feature map is $1/4$ scaled compared to the original image. The features are then up-sampled with a pixel shuffle layer to make the size of the feature map match the input. We make this modification because many endpoints in floor plans are close to each other. Using bins instead of pixels can result in a significant drop in the recall rate (Section \ref{sec: exp}). Hence, the junction offset module presented in \cite{zhou2019end} and \cite{xue2020holistically} are removed. The neural network only predicts the junction likelihood $\hat{J}'(p)$, that for each pixel we have

\begin{equation}
    \hat{J}'(p) =
    \begin{cases}
        \hat{J}(p) & \hat{J}(p) = max_{p' \in N(p)} \hat{J}(p')\\
        0 & \text{otherwise}
    \end{cases}
\end{equation}
where a non-maximum suppression is applied.

\subsection{Line Segment Classification with Graph} \label{subsec3: graph}

We use an attention-based graph network for line segment classification. Attention mechanisms are now widely used in sequence-based tasks and have the advantage of amplifying the most important parts of graphs. This has proven to be useful for many tasks. To learn the relationships between line segments, the inputs to the graph neural network, similar to the definition of a dual graph, is defined on an undirected graph $\mathcal{G}'=(\mathcal{V},\mathcal{E})$ where $v\in\mathcal{V}=\{1,2,...,n\}$ represents the $v$-th line segment, which type is represented by $c'_{v}$. If $e_{v_0,v_1}=1$ where $e_{v_0,v_1}\in\mathcal{E}$ and $v_0,v_1\in\mathcal{V},v_0 \neq v_1$, the $v_0$-th and $v_1$-th line segment has a common junction. To clarify, the word "junction" and "line segment" correspond to $\mathcal{P}$ and $\mathcal{A}$ in the objective graph $\mathcal{G}$, respectively; while the word "node" and "edge" correspond to $\mathcal{V}$ and $\mathcal{E}$ in the intermediate graph $\mathcal{G}'$, respectively.

\subsubsection{Find potential nodes}\label{subsec331} Averagely, a floor plan graph contains 50-100 junctions, which means a fully connected graph would involve up to 5,000 line segments. As for the intermediate graph $\mathcal{G}'$, that is thousands of nodes and millions of edges. Hence, we provide two node suppression strategies.

\textbf{Non-shortest suppression (NSS)} Similar to Non-maximum Suppression (NMS), NSS selects line segments out of a fully connected graph by the angles and lengths. If two line segments have a common point, and the angle between is less than $\mathcal{D}_{NSS}$, the longer line segment would be removed. Note that $\mathcal{D}_{NSS}$ is a dynamic threshold depending on the length and the direction of the longer line segment that

\begin{equation}
    \mathcal{D}_{NSS} =
    \begin{cases}
        2^{\circ} & \text{the line is ``potential"}\\
        22.5^{\circ} & \text{otherwise}
    \end{cases}
\end{equation}
In this paper, the line is ``potential" if its length is less than 20 pixels or

\begin{equation}
    \min(\theta_0, \theta_1, \theta_2, \theta_3) < \frac{200}{l} + 2
\end{equation}
where $\theta_0$, $\theta_1$, $\theta_2$, and $\theta_3$ are the angles (in degrees) between the line segment and the vector $(0, 1)$, $(0, -1)$, $(1, 0)$, $(-1, 0)$, respectively. Here, $l$ represents the length of the line segment (in pixels).

\textbf{Non-diagonal suppression (NDS)} NDS is more aggressive comparing to NSS that the line is discarded if it is not ``potential". Note that not all line segments in floor plans are horizontal or vertical lines. The inclined walls are usually longer due to aesthetic and architectural reasons \cite{Lv_2021_CVPR}, so that the line segments of the convex hull are also added to the set of potential line segments regardless of the above suppression.

\subsubsection{Embeddings}\label{subsec332}

\begin{figure}[!t]
    \centering
    \includegraphics[width=.99\linewidth]{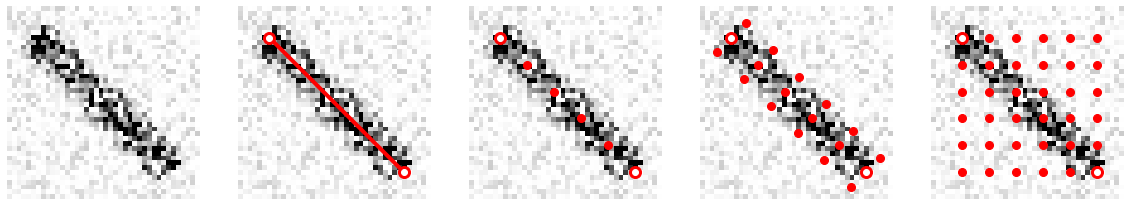}
    \caption{Visual comparison of different pooling methods. From left to right, 1) feature map; 2) the proposed line; 3) line of interest (LoI); 4) rotated region of interest (RRoI); 5) region of interest (RoI). The red dots in the 3rd to 5th picture represent the sampling points given by each pooling method.}\label{Fig:loi_rroi_roi}
\end{figure}

The primitive edge embedding consists of the normalized 2-dimensional coordinate values of the junction. The range of which is bounded by $[-1, 1)$. The primitive node embedding includes: 1) basic information about the line segment, and 2) line features extracted from the second feature map. The vectors are concatenated to form the node embedding.

The basic information of line segments contains the normalized coordinates of the midpoint and two endpoints of the line segment, the length of the line segment, and the absolute cosine value of the angle between the line segment and the horizontal axis. Note the order of the endpoints should not affect the result, so that the endpoints are randomly switched in our implementation.

To extract the feature vector of each line segment, we introduce Rotated Region of Interest (RRoI) Pooling. It conceptually combines the advantages of LoI pooling and RoI Align. In wireframe detection, each ground truth line segment is usually in a position where the color gradient changes drastically so that the width is usually narrow. However, line segments in a floor plan are usually thick or painted with unique textures to represent special structures such as sliding doors and load-bearing walls. Hence, in addition to selecting sampling points along the direction of the line segment, RRoI also selects sampling points along the normal of the line segment as shown in Figure \ref{Fig:loi_rroi_roi}. The set of distances along the normal used in our model is $\{-1, 0, 1\}$, and the number of points along the line is 32. A 2D max-pooling layer with a kernel size of (2,3) is used to reduce the shape of the feature from 32 by 3 to 16 by 1.

\subsubsection{Connection Classification}\label{subsec333}

We adopt an attention based Graph Neural Network (GNN) similar to the design of Gated Attention Network (GaAN) \cite{zhang2018gaan} as our classification model to capture the relationship between nodes and classify the type of each node:

\begin{equation}
\mathbf{x}_{m,i}=\mathrm{FC}_{\theta_{o}}\bigl(\mathbf{x}_{m-1,i} \oplus \mathop{\|}\limits^{K}_{k=1} \sum_{j \in \mathcal{N}_{i}} w_{i, j}^{(k)} \circ \mathrm{FC}_{\theta_{v}^{(k)}}^{h}(\mathbf{z}_{j})\bigr)
\end{equation}
Here, $\mathbf{x}_{m,i}$ is the vector of node $i$ at the $m$-th iteration. $\mathcal{N}_{i}$ represents node $i$'s neighbours, and $\mathbf{z}$ is the reference vector of a neighbour node. $K$ is the number of heads, and both $\oplus$ and $\|$ are the concatenation operation. $\circ$ represents element-wise multiplication. FC means fully connected layers, and $\theta$ represents the corresponding parameters. The formulation of the channel-wise attention between node $i$ and its neighbour $j$ is

\begin{equation}
w_{i, j, c}^{(k)}=\frac{\exp \left(\phi_{w,c}^{(k)}\left(\mathbf{x}_{i}, \mathbf{z}_{j}, \mathbf{e}_{i,j}\right)\right)}{\sum_{l=1}^{\left|\mathcal{N}_{i}\right|} \exp \left(\phi_{w,c}^{(k)}\left(\mathbf{x}_{i}, \mathbf{z}_{l}, \mathbf{e}_{i,l}\right)\right)}
\end{equation}
where $c$ represents $c$-th channel, and $\mathbf{e}_{i,j}$ is the feature of the edge from $i$ to $j$. The dot product attention is replaced by fully connected layers to aggregate information of edges:

\begin{equation}
\phi_{w}^{(k)}(\mathbf{x}, \mathbf{z}, \mathbf{e})=\mathrm{FC}_{\theta_{w}^{(k)}} \left(\psi_{w}^{(k)}(\mathbf{x}, \mathbf{z}, \mathbf{e})\right)
\end{equation}
Here, $\psi_{w}^{(k)}(\mathbf{x}, \mathbf{z}, \mathbf{e})$ concatenates the projected features:

\begin{equation}
\psi_{w}^{(k)}(\mathbf{x}, \mathbf{z}, \mathbf{e})=\mathrm{FC}_{\theta_{x a}^{(k)}}(\mathbf{x}) \oplus \mathrm{FC}_{\theta_{z a}^{(k)}}(\mathbf{z}) \oplus \mathrm{FC}_{\theta_{e}^{(k)}}(\mathbf{e})
\end{equation}
The output vector of node $i$ is

\begin{equation}
\mathbf{y}_{i}=\sigma(\mathbf{x}_{M,i})
\end{equation}
where $M$ is the depth of the GNN, and $\sigma(\cdot)$ is the sigmoid function which determines the likelihood of each line segment category.

\subsection{Multi-task Learning} \label{subsec3: loss}

We use the segmentation map of the structural elements (walls, door, and windows) to supervise the intermediate layers of hourglass modules. The proposed model is trained with the following loss function:

\begin{equation}
\mathbb{L}=\mathbb{L}_{Hourglass}+\mathbb{L}_{Junc}+\mathbb{L}_{Graph}
\end{equation}
where all three losses are binary cross entropy loss. In our experiments, $\mathbb{L}_{Graph}$ is not added for the first 10,000 steps. The category of a node in the ground truth of the intermediate graph $\mathcal{G}'$ is regarded as one of the above structural elements if the $L_2$ distance between the node and any of these ground truths is less than $d_{max}=25$.



\begin{figure}[!t]
    \centering
    \includegraphics[width=.99\linewidth]{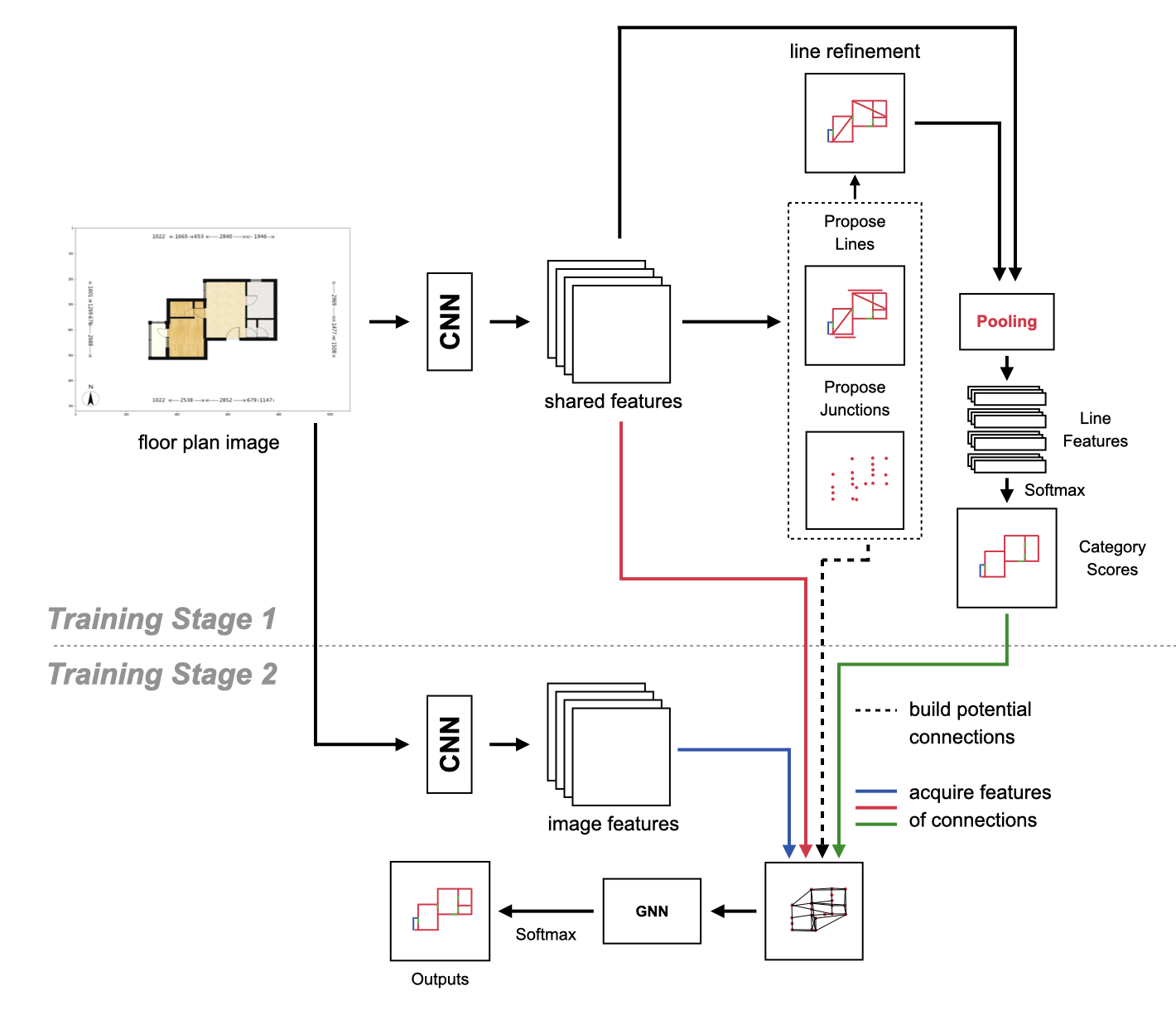}
    \caption{An overview of the model structure for the integratable module paradigm. Stage 1 and stage 2 are trained independently.}\label{Fig:model_structure_pp}
\end{figure}

\section{Experiments} \label{sec: exp}

\subsection{Baselines}\label{subsec41}

\textbf{Modified HAWP} We choose the previous state-of-the-art model in wireframe parsing as one of our baseline approaches, yet the model is not designed for multi-class line segment detection. Thus, a few modifications are made: 1) The fully connected layers are no longer projecting LoI features to scores but to categories. 2) The junction detection module is aligned with our method described in Section \ref{subsec3: junc}. 3) The RRoI line feature extraction module is introduced to replace LoI. The effect of each modification is discussed in Section \ref{subsec44}. As in the original paper, the binary cross entropy loss is used to train the models.

\begin{table}[!b]
    \centering
    \caption{Quantitative evaluation of junction detection. The best results are in bold texts, and the second best results are emphasized with underlines. We report results averaged over 3 experiments. The PR curves show in Figure \ref{Fig:pr_junction} are drawn based on the first experiment. Please refer to the supplementary material for the remaining curves. Same for Table \ref{tab:line_seg_eval}, \ref{tab:line_seg_eval_1}, \ref{tab:prior_eval}, and Figure \ref{Fig:line_pr_line}, \ref{Fig:line_pr_line_1_2}, \ref{Fig:prior_pr_line}.}
    \setlength{\tabcolsep}{1em}
    \resizebox{0.99\linewidth}{!}{
    \begin{tabular}{l c c c }
    \toprule
        & sAP$^{2}_J$ & sAP$^{4}_J$ & sAP$^{8}_J$ \\
        \midrule
        HAWP-M & $61.78 \textcolor{blue}{\pm 0.24}$ & $69.72 \textcolor{blue}{\pm 0.19}$ & $72.79 \textcolor{blue}{\pm 0.21}$ \\
        \cmidrule(lr){2-4}
        HAWP-M+, ksize=7 & $63.81 \textcolor{blue}{\pm 0.30}$ & $79.16 \textcolor{blue}{\pm 0.16}$ & $81.55 \textcolor{blue}{\pm 0.10}$ \\
        \cmidrule(lr){2-4}
        HAWP-M+, ksize=5 & $65.43 \textcolor{blue}{\pm 0.43}$ & $81.19 \textcolor{blue}{\pm 0.22}$ & $83.57 \textcolor{blue}{\pm 0.11}$ \\
        \cmidrule(lr){2-4}
        HAWP-M+, ksize=3 & $65.03 \textcolor{blue}{\pm 0.32}$ & $81.76 \textcolor{blue}{\pm 0.2}$ & $84.46 \textcolor{blue}{\pm 0.12}$ \\
        \cmidrule(lr){2-4}
        HAWP-M* & $65.08 \textcolor{blue}{\pm 0.51}$ & $81.86 \textcolor{blue}{\pm 0.25}$ & $84.60 \textcolor{blue}{\pm 0.08}$ \\
        \midrule
        GLSP (NDS) & $\textbf{76.64} \textcolor{blue}{\pm 0.16}$ & $\textbf{87.97} \textcolor{blue}{\pm 0.07}$ & $\textbf{89.38} \textcolor{blue}{\pm 0.02}$ \\
        \cmidrule(lr){2-4}
        GLSP (NDS + SF) & $\underline{75.98} \textcolor{blue}{\pm 0.22}$ & $\underline{87.72} \textcolor{blue}{\pm 0.06}$ & $\underline{89.28} \textcolor{blue}{\pm 0.04}$ \\
        \cmidrule(lr){2-4}
        GLSP (NSS + SF) & $75.08 \textcolor{blue}{\pm 0.27}$ & $87.54 \textcolor{blue}{\pm 0.18}$ & $89.23 \textcolor{blue}{\pm 0.07}$ \\
    \toprule
    \end{tabular}
    }
    \label{tab:junc_detect_eval}
\end{table}

\begin{figure}[!b]
    \centering
    \includegraphics[width=.99\linewidth]{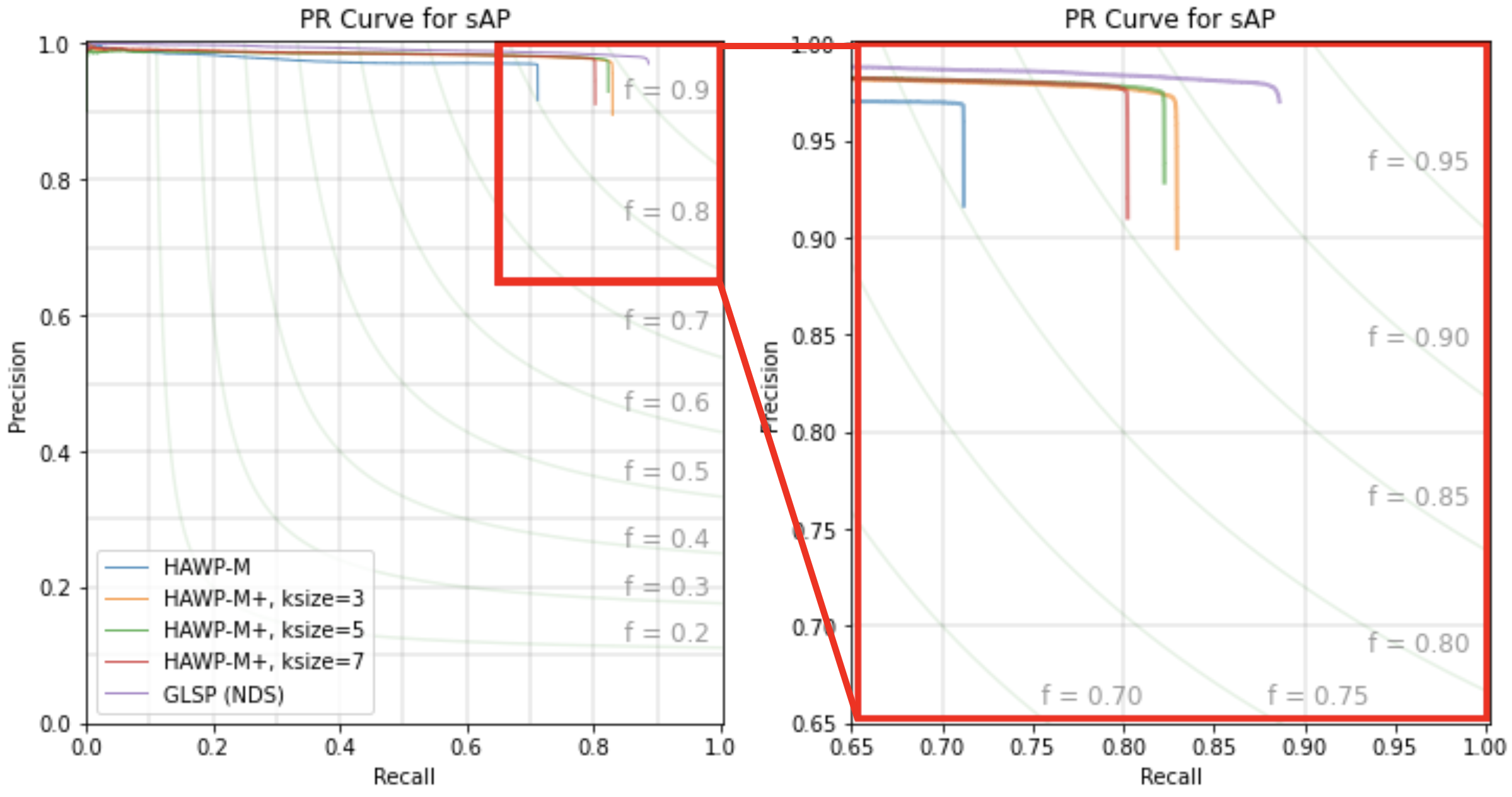}
    \caption{Precision-Recall (PR) curves of junction detections.}\label{Fig:pr_junction}
\end{figure}

\begin{table*}[!t]
    \centering
    \caption{Quantitative evaluation of line segment detection.}
    \setlength{\tabcolsep}{1em}
    \resizebox{0.99\textwidth}{!}{
    \begin{tabular}{l c c c c c c }
    \toprule
        & msAP$^{8}$ & msAP$^{16}$ & msAP$^{32}$ & sAP$^{8}_N$ & sAP$^{16}_N$ & sAP$^{32}_N$ \\
        \midrule
        HAWP-M & $55.45 \textcolor{blue}{\pm 0.37}$ & $57.07 \textcolor{blue}{\pm 0.35}$ & $57.48 \textcolor{blue}{\pm 0.35}$ & $53.89 \textcolor{blue}{\pm 0.22}$ & $54.93 \textcolor{blue}{\pm 0.19}$ & $55.22 \textcolor{blue}{\pm 0.18}$ \\
        \cmidrule(lr){2-7}
        HAWP-M+ & $70.72 \textcolor{blue}{\pm 0.32}$ & $72.69 \textcolor{blue}{\pm 0.31}$ & $73.51 \textcolor{blue}{\pm 0.3}$ & $69.66 \textcolor{blue}{\pm 0.17}$ & $71.68 \textcolor{blue}{\pm 0.15}$ & $72.49 \textcolor{blue}{\pm 0.14}$ \\
        \cmidrule(lr){2-7}
        HAWP-M* & $71.12 \textcolor{blue}{\pm 0.32}$ & $73.33 \textcolor{blue}{\pm 0.34}$ & $74.30 \textcolor{blue}{\pm 0.30}$ & $69.76 \textcolor{blue}{\pm 0.18}$ & $71.95 \textcolor{blue}{\pm 0.17}$ & $72.86 \textcolor{blue}{\pm 0.18}$ \\
        \cmidrule(lr){2-7}
        HAWP-M* + GNN & $\underline{85.86} \textcolor{blue}{\pm 0.35}$ & $\underline{91.50} \textcolor{blue}{\pm 0.16}$ & $\underline{92.50} \textcolor{blue}{\pm 0.04}$ & $\underline{81.38} \textcolor{blue}{\pm 0.27}$ & $\underline{85.25} \textcolor{blue}{\pm 0.13}$ & $\underline{85.87} \textcolor{blue}{\pm 0.06}$ \\
        \cmidrule(lr){2-7}
        GLSP & $\textbf{90.28} \textcolor{blue}{\pm 0.36}$ & $\textbf{93.59} \textcolor{blue}{\pm 0.12}$ & $\textbf{94.23} \textcolor{blue}{\pm 0.05}$ & $\textbf{87.19} \textcolor{blue}{\pm 0.22}$ & $\textbf{89.98} \textcolor{blue}{\pm 0.04}$ & $\textbf{90.44} \textcolor{blue}{\pm 0.01}$ \\
    \toprule
    \end{tabular}
    }
    \label{tab:line_seg_eval}
\end{table*}

\begin{figure*}[!t]
    \centering
    \includegraphics[width=.99\linewidth]{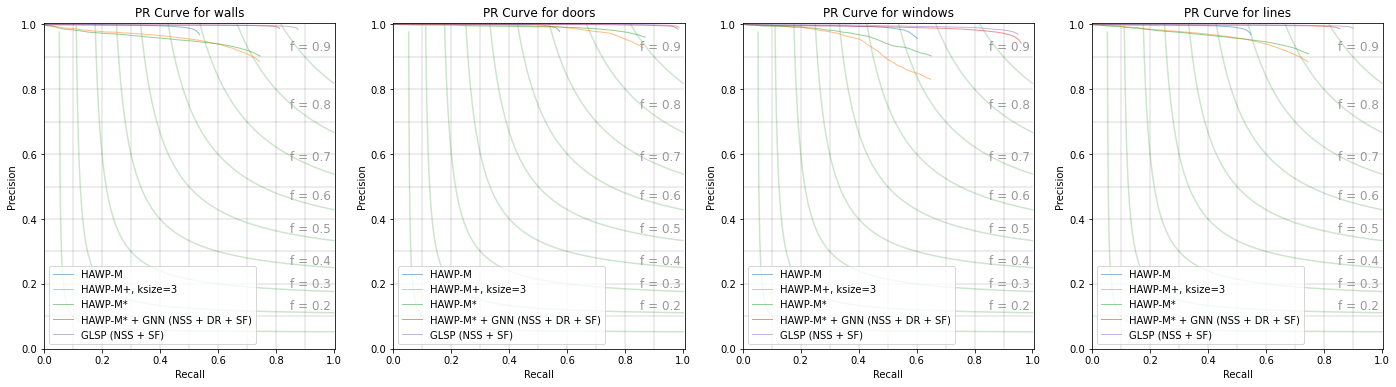}
    \caption{Precision-Recall (PR) curves of line segment detections.}\label{Fig:line_pr_line}
\end{figure*}

\textbf{GLSP as an integratable module} The GLSP model can also be used as an integratable module on a conventional line segment detection algorithm. Here, we choose the modified HAWP as the line segment detection algorithm, and the same techniques described in Section \ref{subsec331} to build the graph. The line segment classification results given by the modified HAWP are added to the features of nodes (the green line in Figure \ref{Fig:model_structure_pp}), and the performance of adding line features extracted from the feature map of the modified HAWP model into the GNN (the red line in Figure \ref{Fig:model_structure_pp}) is also tested in the ablation study. The two modules are trained independently, so the parameters of the modified HAWP would not be updated when training the GNN. The binary cross entropy loss is used to train the GNN.

\subsection{Implementation Details}

To be fair, the hyperparameters will be consistent with the settings in HAWP for the line segment detection and endpoint detection modules if applicable. The input images are padded by replicating the pixel values at the edges of the original image and resized to $512 \times 512$ for both training and evaluation. The Stacked Hourglass backbones mentioned in Section \ref{subsec3: data} and Section \ref{subsec41} are identical. Same as HAWP, the number of stacks, the depth of each module, and the number of blocks are 2, 4, and 1, respectively. The modified HAWPs are trained for 8 epochs with a batch size of 16. The learning rates are $4\times 10^{-4}$ for the first 6 epochs, and $4\times 10^{-5}$ for the last two.

As for GLSPs, the depth of GNN is 4, and the number of heads is 8. The dimensions of outputs for $\mathrm{FC}_{\theta_{x a}}$, $\mathrm{FC}_{\theta_{z a}}$, $\mathrm{FC}_{\theta_{w}}$, $\mathrm{FC}_{\theta_{v}}$, and $\mathrm{FC}_{\theta_{o}}$ are 128, but equals 4 for the final $\mathrm{FC}_{\theta_{o}}$. The dimensions of outputs for $\mathrm{FC}_{\theta_{e}}$ is 64. No matter it is end-to-end or served as an integratable module, the models are trained for 2 epochs, where the learning rate is $2\times 10^{-4}$ for the first epoch, and $2\times 10^{-5}$ for the other. The batch size equals 8 if GLSP is an integratable module and NDS is used as the suppression strategy, and 4 otherwise.

All models are optimized by the ADAM optimizer \cite{kingma2014adam} setting the weight decay to $1\times10^{-4}$. HAWP-M, HAWP-M+, and HAWP-M* in the following experiments represent the HAWP with the first, the first two, and all modifications in Section \ref{subsec41}. The kernel size for NMS is 3 if not mentioned. In Table \ref{tab:line_seg_eval}, the model "HAWP-M* + GNN" does not use line segment feature extracted from the feature map of the modified HAWP (the red line in figure \ref{Fig:model_structure_pp}), which effect are evaluated in the ablation study section.

\begin{table*}[!t]
    \centering
    \caption{The ablation study of different designs. If NSS is not marked, NDS is used as the suppression strategy.}
    \setlength{\tabcolsep}{1em}
    \resizebox{0.99\textwidth}{!}{
    \begin{tabular}{c c c c c c c c c c c }
    \toprule
        BI & PR & DR & SF & NSS & msAP$^{8}$ & msAP$^{16}$ & msAP$^{32}$ & sAP$^{8}_N$ & sAP$^{16}_N$ & sAP$^{32}_N$ \\
        \cmidrule(lr){1-5} \cmidrule(lr){6-11}
        \multicolumn{4}{l}{HAWP-M* + GNN} \\
        \midrule
        $\bigcirc$ & $\bigcirc$ & & & & $79.66 \textcolor{blue}{\pm 0.35}$ & $85.07 \textcolor{blue}{\pm 0.08}$ & $85.96 \textcolor{blue}{\pm 0.07}$ & $80.19 \textcolor{blue}{\pm 0.29}$ & $84.38 \textcolor{blue}{\pm 0.12}$ & $85.04 \textcolor{blue}{\pm 0.06}$ \\
        \cmidrule(lr){6-11}
        $\bigcirc$ & & $\bigcirc$ & & & $85.16 \textcolor{blue}{\pm 0.32}$ & $90.24 \textcolor{blue}{\pm 0.14}$ & $91.13 \textcolor{blue}{\pm 0.04}$ & $81.03 \textcolor{blue}{\pm 0.20}$ & $84.84 \textcolor{blue}{\pm 0.11}$ & $85.44 \textcolor{blue}{\pm 0.05}$ \\
        \cmidrule(lr){6-11}
        $\bigcirc$ & $\bigcirc$ & $\bigcirc$ & & & $85.37 \textcolor{blue}{\pm 0.32}$ & $90.47 \textcolor{blue}{\pm 0.16}$ & $91.36 \textcolor{blue}{\pm 0.06}$ & $\underline{81.23} \textcolor{blue}{\pm 0.20}$ & $84.91 \textcolor{blue}{\pm 0.11}$ & $85.50 \textcolor{blue}{\pm 0.05}$ \\
        \cmidrule(lr){6-11}
        $\bigcirc$ & & $\bigcirc$ & $\bigcirc$ & & $\underline{85.72} \textcolor{blue}{\pm 0.27}$ & $\underline{91.30} \textcolor{blue}{\pm 0.14}$ & $\underline{92.26} \textcolor{blue}{\pm 0.03}$ & $81.17 \textcolor{blue}{\pm 0.21}$ & $\underline{85.13} \textcolor{blue}{\pm 0.11}$ & $\underline{85.75} \textcolor{blue}{\pm 0.05}$ \\
        \cmidrule(lr){6-11}
        $\bigcirc$ & $\bigcirc$ & $\bigcirc$ & $\bigcirc$ & & $85.11 \textcolor{blue}{\pm 0.29}$ & $90.70 \textcolor{blue}{\pm 0.11}$ & $91.66 \textcolor{blue}{\pm 0.02}$ & $80.78 \textcolor{blue}{\pm 0.28}$ & $84.91 \textcolor{blue}{\pm 0.12}$ & $85.56 \textcolor{blue}{\pm 0.05}$ \\
        \cmidrule(lr){6-11}
        $\bigcirc$ & & $\bigcirc$ & $\bigcirc$ & $\bigcirc$ & $\textbf{85.86} \textcolor{blue}{\pm 0.35}$ & $\textbf{91.50} \textcolor{blue}{\pm 0.16}$ & $\textbf{92.5} \textcolor{blue}{\pm 0.04}$ & $\textbf{81.38} \textcolor{blue}{\pm 0.27}$ & $\textbf{85.25} \textcolor{blue}{\pm 0.13}$ & $\textbf{85.87} \textcolor{blue}{\pm 0.06}$ \\
        \midrule
        \multicolumn{4}{l}{GLSP} \\
        \midrule
        $\bigcirc$ & & & & & $86.00 \textcolor{blue}{\pm 0.20}$ & $89.10 \textcolor{blue}{\pm 0.16}$ & $89.72 \textcolor{blue}{\pm 0.07}$ & $\underline{86.65} \textcolor{blue}{\pm 0.05}$ & $89.29 \textcolor{blue}{\pm 0.03}$ & $89.76 \textcolor{blue}{\pm 0.03}$ \\
        \cmidrule(lr){6-11}
        $\bigcirc$ & & & $\bigcirc$ & & $\underline{89.13} \textcolor{blue}{\pm 0.14}$ & $\underline{92.25} \textcolor{blue}{\pm 0.03}$ & $\underline{92.87} \textcolor{blue}{\pm 0.07}$ & $86.56 \textcolor{blue}{\pm 0.05}$ & $\underline{89.38} \textcolor{blue}{\pm 0.05}$ & $\underline{89.84} \textcolor{blue}{\pm 0.07}$ \\
        \cmidrule(lr){6-11}
        $\bigcirc$ & & & $\bigcirc$ & $\bigcirc$ & $\textbf{90.28} \textcolor{blue}{\pm 0.36}$ & $\textbf{93.59} \textcolor{blue}{\pm 0.12}$ & $\textbf{94.23} \textcolor{blue}{\pm 0.05}$ & $\textbf{87.19} \textcolor{blue}{\pm 0.22}$ & $\textbf{89.98} \textcolor{blue}{\pm 0.04}$ & $\textbf{90.44} \textcolor{blue}{\pm 0.01}$ \\
    \toprule
    \end{tabular}
    }
    \label{tab:line_seg_eval_1}
\end{table*}

\begin{figure*}[!t]
    \centering
    \begin{subfigure}[c]{.99\linewidth}
        \centering
        \includegraphics[width=.99\linewidth]{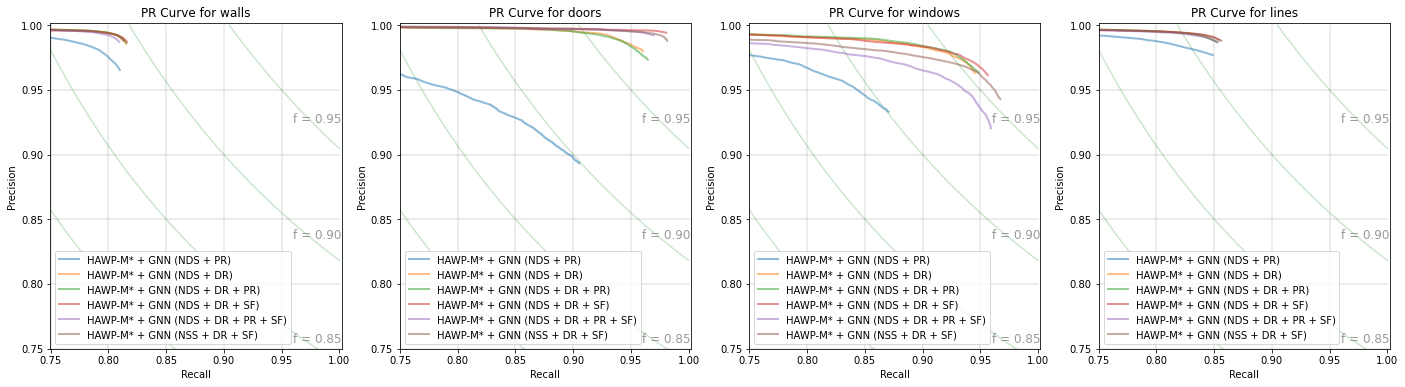}
        \caption{}\label{Fig:line_pr_line_1}
    \end{subfigure}
    \hfill
    \begin{subfigure}[c]{.99\linewidth}
        \centering
        \includegraphics[width=.99\linewidth]{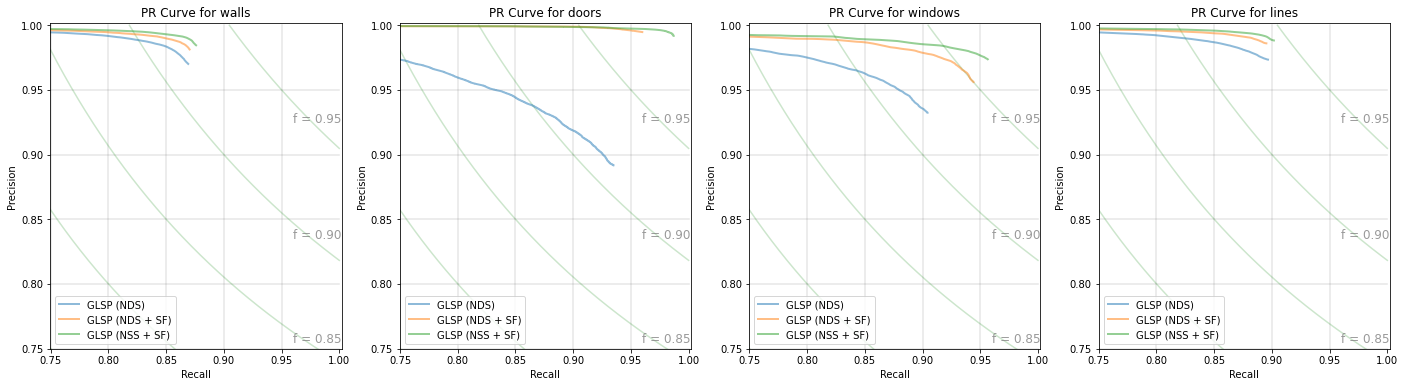}
        \caption{}\label{Fig:line_pr_line_2}
    \end{subfigure}
    \caption{Precision-Recall (PR) curves of (a) GLSP as integratable modules, and (b) end-to-end GLSPs.}\label{Fig:line_pr_line_1_2}
\end{figure*}

\subsection{Evaluation Metrics}

We follow the definition of Structural Average Precision ($\text{sAP}$) used in \cite{zhou2019end} and \cite{xue2020holistically}, whereas the results are evaluated on the ground truths with resolutions of $512 \times 512$ rather than $128 \times 128$ in \cite{xue2020holistically}. Please refer to the supplementary material for the results of $\text{sAP}$ for each class (wall, door, and window). $\text{sAP}_{N}$ represents Structural Average Precision without considering the class of line segments. The mean of $\text{sAP}$ values over different categories is denoted as $\text{msAP}$. We set the threshold of $L_2$ distance $\vartheta_L$ to 8, 16, 32, and denote the results as $\text{msAP}^{\vartheta_L}$ and $\text{sAP}_{N}^{\vartheta_L}$. The vectorized junction AP ($\text{sAP}_{J}$) is designed in a similar way, where the threshold $\vartheta_J \in \{2, 4, 8\}$.

\subsection{Results and Analysis}\label{subsec44}

Table \ref{tab:line_seg_eval} and Figure \ref{Fig:line_pr_line} shows the performance of line segment detection for baseline models and GLSP. The modifications made to HAWP improve its performance, whereas using GLSP as an integratable module and the end-to-end GLSP is a better choice for both junction detection and line segment detection. In Section \ref{subsec3: junc}, we argue that using bins instead of pixels can result in a significant drop in the recall rate of junction detection, which is verified in Table \ref{tab:junc_detect_eval} and Figure \ref{Fig:pr_junction}. By comparing the accuracy of different models, it is not difficult to infer that the accuracy of junction detection can influence the accuracy of line segment detection to some extent. We also adjust the kernel size of NMS, and it can be seen from the table that a larger range of NMS does harm junction detection. GLSP predicts the junctions much better than HAWP-M*, and we suggest the reason may be that the feature map is only used to detect junctions and does not need to be shared with the module of line segment proposal. Please refer to the supplementary material for qualitative examples from the models mentioned above.

\subsection{Ablation Study}

We conducted ablation studies from two aspects: 1) the effect of assigning different features to nodes when building the intermediate graph $\mathcal{G}'$, and 2) adding prior knowledge or limitations to the training process.

Table \ref{tab:line_seg_eval_1} summarizes the comparisons. BI, DR, PR, and SF represent assigning the basic information of the line segments, assigning line segment detection results (the green line in Figure \ref{Fig:model_structure_pp}), assigning the pooling results of line segment detection network (the red line in Figure \ref{Fig:model_structure_pp}), and assigning line features extracted from the second image feature extraction networks (the blue line in Figure \ref{Fig:model_structure_1} and \ref{Fig:model_structure_pp}) to the nodes of the intermediate graph $\mathcal{G}'$, respectively.

\begin{table*}[!t]
    \centering
    \caption{Quantitative evaluation of line segment detection with prior knowledge.}
    \setlength{\tabcolsep}{1em}
    \resizebox{0.99\textwidth}{!}{
    \begin{tabular}{l c c c c c c c c }
    \toprule
        & $N_r$ & $N_R$ & msAP$^{8}$ & msAP$^{16}$ & msAP$^{32}$ & sAP$^{8}_N$ & sAP$^{16}_N$ & sAP$^{32}_N$ \\
        \cmidrule(lr){1-1} \cmidrule(lr){2-3} \cmidrule(lr){4-9}
        GLSP (NDS + SF) & $5.22 \textcolor{blue}{\pm 0.07}$ & $5.17 \textcolor{blue}{\pm 0.06}$ & $89.13 \textcolor{blue}{\pm 0.14}$ & $92.25 \textcolor{blue}{\pm 0.03}$ & $92.87 \textcolor{blue}{\pm 0.07}$ & $86.56 \textcolor{blue}{\pm 0.05}$ & $89.38 \textcolor{blue}{\pm 0.05}$ & $89.84 \textcolor{blue}{\pm 0.07}$ \\
        \cmidrule(lr){2-3} \cmidrule(lr){4-9}
        GLSP (NDS + SF + PK) & $5.40 \textcolor{blue}{\pm 0.06}$ & $5.35 \textcolor{blue}{\pm 0.06}$ & $89.11 \textcolor{blue}{\pm 0.29}$ & $92.53 \textcolor{blue}{\pm 0.08}$ & $93.19 \textcolor{blue}{\pm 0.03}$ & $86.31 \textcolor{blue}{\pm 0.19}$ & $89.14 \textcolor{blue}{\pm 0.03}$ & $89.63 \textcolor{blue}{\pm 0.01}$ \\
        \cmidrule(lr){2-3} \cmidrule(lr){4-9}
        GLSP (NSS + SF) & $\underline{5.49} \textcolor{blue}{\pm 0.06}$ & $\underline{5.46} \textcolor{blue}{\pm 0.06}$ & $\underline{90.28} \textcolor{blue}{\pm 0.36}$ & $\underline{93.59} \textcolor{blue}{\pm 0.12}$ & $\underline{94.23} \textcolor{blue}{\pm 0.05}$ & $\textbf{87.19} \textcolor{blue}{\pm 0.22}$ & $\textbf{89.98} \textcolor{blue}{\pm 0.04}$ & $\textbf{90.44} \textcolor{blue}{\pm 0.01}$ \\
        \cmidrule(lr){2-3} \cmidrule(lr){4-9}
        GLSP (NSS + SF + PK) & $\textbf{5.59} \textcolor{blue}{\pm 0.06}$ & $\textbf{5.56} \textcolor{blue}{\pm 0.06}$ & $\textbf{90.43} \textcolor{blue}{\pm 0.21}$ & $\textbf{93.88} \textcolor{blue}{\pm 0.10}$ & $\textbf{94.55} \textcolor{blue}{\pm 0.04}$ & $\underline{87.03} \textcolor{blue}{\pm 0.14}$ & $\underline{89.95} \textcolor{blue}{\pm 0.04}$ & $\underline{90.42} \textcolor{blue}{\pm 0.01}$ \\
    \toprule
    \end{tabular}
    }
    \label{tab:prior_eval}
\end{table*}

\begin{figure*}[!t]
    \centering
    \includegraphics[width=.99\linewidth]{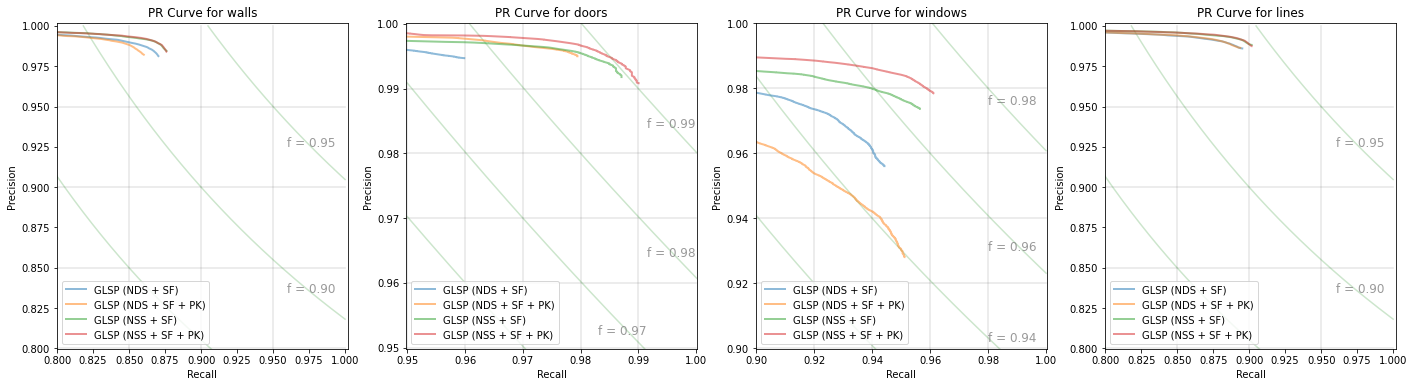}
    \caption{Precision-Recall (PR) curves of line segment detection with prior knowledge. To clarify the difference between curves, the scales of the figures are not identical.}\label{Fig:prior_pr_line}
\end{figure*}

If the GLSP is used as an integratable module, adding the line segment classification results is much better than adding line features extracted from the feature map of the modified HAWP model. The latter can also have negative effects in some cases. Therefore, the model "HAWP-M* + GNN" in Table \ref{tab:line_seg_eval} does not adopt the feature. The second feature map can slightly improve the performance of both paradigms. NSS may be a better choice compared to NDS, but the number of nodes in $\mathcal{G}'$ of NSS is approximately 4 times that of NDS.

Another problem we would like to discuss is adding prior knowledge or limitations to the training process. Here, we introduce two possible prior knowledge to the loss function after the category $c'_{v_i}$ for node $v_i$ is predicted: 1) the graph loss for $v_i$ is doubled if the line segment it represented is intersected with node $v_j$'s corresponding line segment, where $v_i \ne v_j$, while both are classified as meaningful line segments (a wall, a door, or a window), and 2) the graph loss for a node is doubled if a sequence of nodes $K$ can be found that a) $v_i \in K$, b) the line segments represented by the nodes in the sequence are connected end-to-end, and c)$\forall k \in K$, $c'_{k} \in \{\text{wall}, \text{window}\}$, which means the area is not connected to other parts of the floor plan. The losses are introduced after the model has been trained for 20,000 steps. We use PK to represent whether the loss of prior knowledge is used, and $N_r$ and $N_R$ to represent the number of enclosed rooms, and the number of enclosed rooms with doors, respectively. As shown in Table \ref{tab:prior_eval} and Figure \ref{Fig:prior_pr_line}, the strategy improves the number of valid rooms in the final results, but slightly reduces the accuracy of the line segment detection.

\section{Conclusion} \label{sec: conclusion}

In this paper, we present GLSP, a line segment detection algorithm based on Graph Attention Network. The proposed model can be used as an end-to-end algorithm or an integratable module on existing line segment detection models. The former one preform better than the latter paradigm on our open-source floor plan dataset LRFP. Our proposed methods are capable to output vectorized results of junctions and line segments, which may reduce the amount of computation for post-processing works when reconstructing editable floor plans from images.


{\small
\bibliographystyle{ieee_fullname}
\bibliography{mybib}
}

\newpage~

\begin{appendices}
\begin{figure}[b]
    \centering
    \includegraphics[width=.9\linewidth]{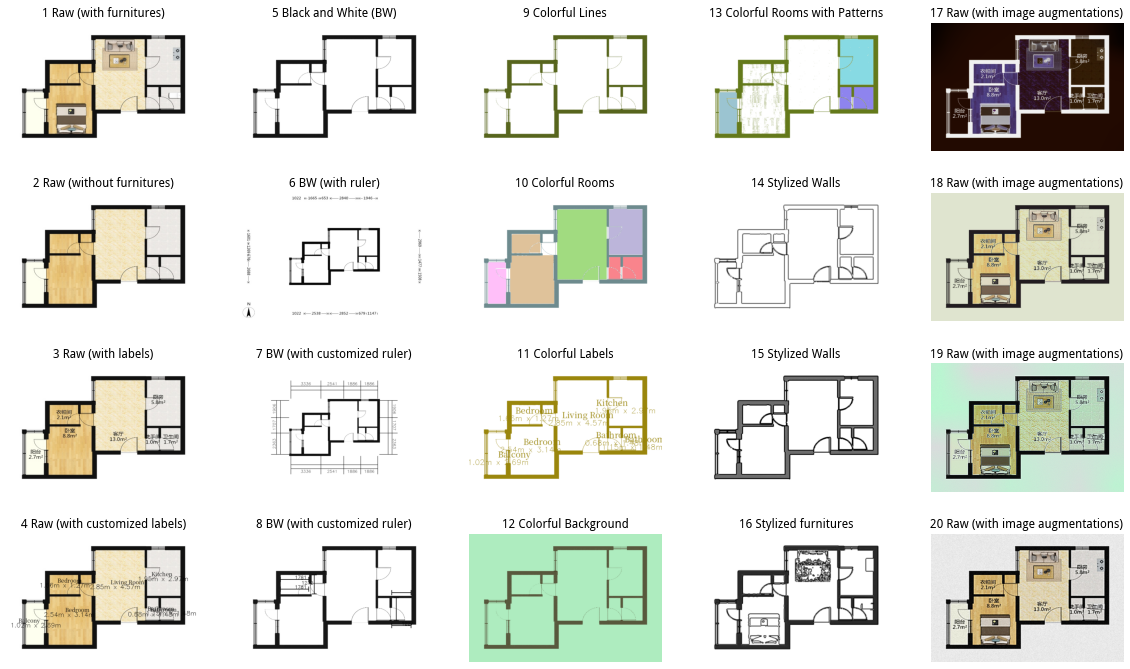}
    \caption{(1)-(4) raw images of four different Levels of Details (LOD); (5) remove the furniture items and patterns; (6) ruler type I; (7) ruler type II; (8) ruler type III; (9) change the color of lines; (10) change the color of lines and rooms; (11) change the color of rooms labels; (12) change the color of the image background; (13) change the color of lines, rooms, and patterns; (14) use hollow lines to represent walls; (15) fill the lines of walls with customized color; (16) draw the contours of furniture items; (17)-(20) examples of classic image augmentation result.}\label{Fig:type_of_aug_00}
\end{figure}

\section{Augmentation Details}

The success of supervised learning relies heavily on big data to avoid overfitting. If the diversity of input images is limited, the model may have a super-human performance on the training set, while facing astonishing failures when applied to real-world scenarios. In the field of computer vision, it is common to use data augmentation techniques to artificially increase the scale of datasets. This applies to floor plan data as well. As shown in figure \ref{Fig:type_of_aug_00}, we use a series of data augmentation methods (including adjusting the style of lines, rooms, rulers, and texts) to make the floor plans applicable to different countries and regions. To ensure the same images are used during different experiments, we pre-augment each image five times using a random combination of methods shown in figure \ref{Fig:type_of_aug_00} followed by classic image augmentation. We use \textit{imgaug} \cite{imgaug} to perform classic image augmentation. The code is as follows:

\begin{lstlisting}
import imgaug.augmenters as iaa
seq = iaa.Sequential([
    iaa.SomeOf((0, 4), [
        iaa.OneOf([
            iaa.GaussianBlur((0, 2.0)),
            iaa.AverageBlur(k=(2, 3)),
        ]),
        iaa.Sharpen(
            alpha=(0, 1.0),
            lightness=(0.75, 1.5)
        ),
        iaa.AdditiveGaussianNoise(
            loc=0, scale=(0.0, 0.05*255),
            per_channel=0.5
        ),
        iaa.Add((-10, 10), per_channel=0.5),
        iaa.AddToHueAndSaturation((-20, 20)),
        iaa.LinearContrast(
            (0.8, 1.2), per_channel=0.5
        ),
    ], random_order=True),
    iaa.SomeOf((0, 2), [
        iaa.OneOf([
            # invert color channels
            iaa.Sometimes(0.25, iaa.Invert(1.0)),
            # add noise
            iaa.BlendAlphaSimplexNoise(
                foreground=iaa.OneOf([
                    iaa.EdgeDetect(
                        alpha=(0.0, 0.5)
                    ),
                    iaa.DirectedEdgeDetect(
                        alpha=(0.0, 0.5),
                        direction=(0.0, 1.0)
                    ),
                ]),
                upscale_method=[
                    "linear", "cubic"
                ],
                size_px_max=(2,16),
            )
        ]),
        # change the brightness
        iaa.OneOf([
            iaa.Multiply(
                (0.8, 1.2),
                per_channel=0.5
            ),
            iaa.BlendAlphaFrequencyNoise(
                exponent=(-4, 0),
                foreground=iaa.Multiply(
                    (0.8, 1.2),
                    per_channel=True
                )
            )
        ]),
    ], random_order=True),
], random_order=True)

\end{lstlisting}

\pagebreak

\section{Dataset Statistics}

\begin{table}[!htp]
    \centering
    \caption{Number of appearance for each type of lines and rooms. The room types represented by ``Others" in the table are: Garden, stairs, elevators, void, and unrecognized.}
    \setlength{\tabcolsep}{1em}
    \resizebox{0.99\columnwidth}{!}{%
    \begin{tabular}{ l | r || l | r }
    \toprule
        & appearances & & appearances \\
        \toprule
        Wall & 13,289,372 & Bedroom & 663,623 \\
        Window & 2,221,027 & Bathroom & 346,271 \\
        Door & 1,763,580 & Balcony & 329,317 \\
        & & Living Room & 273,854 \\
        & & Kitchen & 272,139 \\
        & & Aisle & 67,735 \\
        & & Multi-functional & 41,175 \\
        & & Dining Room & 31,764 \\
        & & Porch & 15,801 \\
        & & Reading Room & 8,985 \\
        & & Others & 2,871 \\
        \midrule
        \textbf{Total} & 17,552,431 & \textbf{Total} & 2,053,535 \\
    \toprule
    \end{tabular}
    }
    \label{tab: number stats}
\end{table}

Table \ref{tab: number stats} shows the frequency of occurrence of different room types in the dataset. Most of the real estate sold in Chinese major cities are medium-sized apartment houses containing one to three bedrooms. This results in an uneven distribution of different types of rooms. The scales of the floor plans are recorded in mm per pixel. The distribution shows an obvious long-tail effect, where the mean, standard deviation, and median are 21.20, 2.41, and 19.80, respectively. More than 99\% of the floor plans have a scale of less than 30. The minimal value is 18.47, and the maximum is 38.16.

\newpage

\section{Supplementary Experimental Results}

\begin{figure*}[!t]
    \centering
    \includegraphics[width=.99\linewidth]{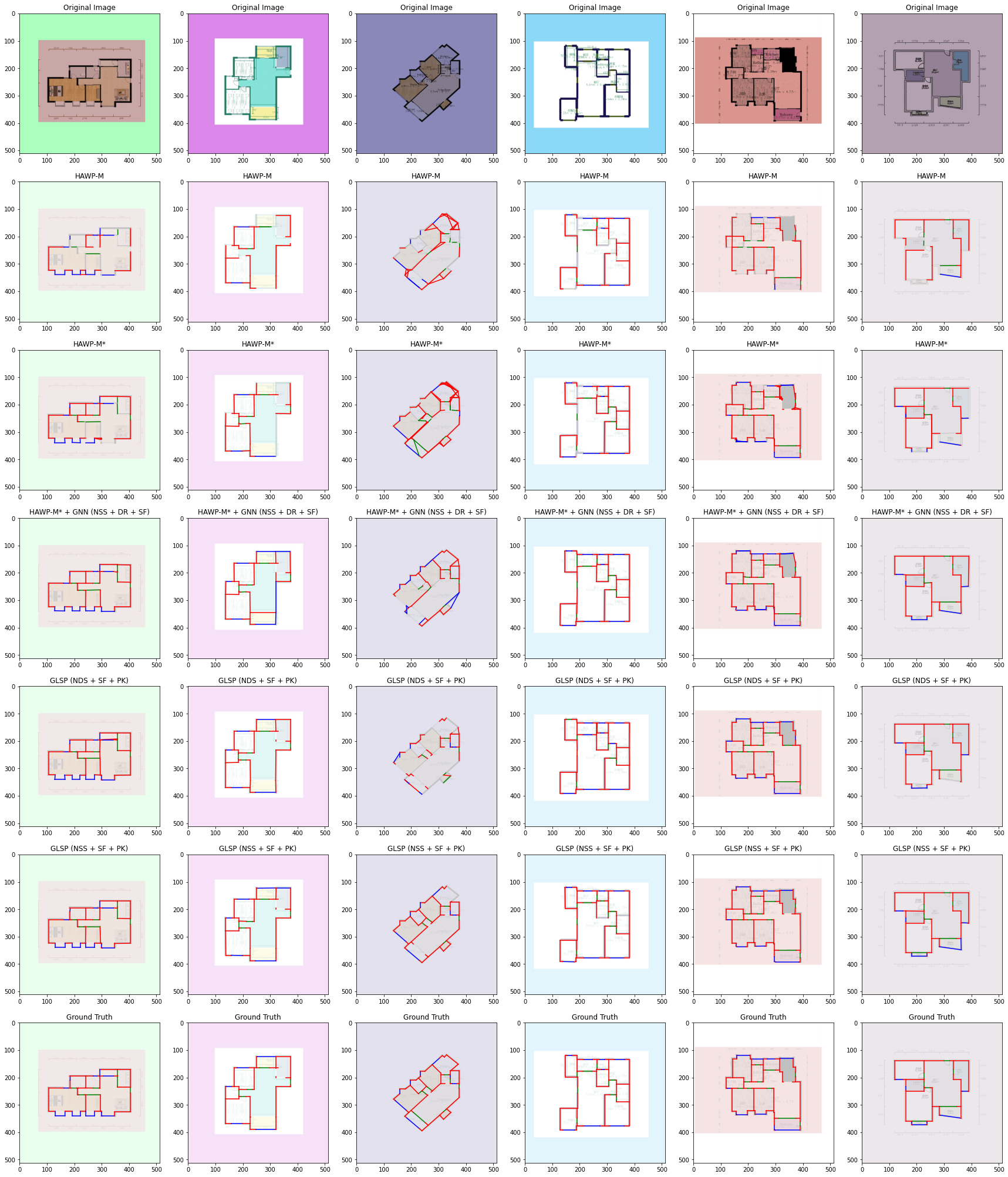}
    \caption{Line segment detection examples on the test set. The walls, doors, and windows in the figures are represented by red, green, and blue line segments, respectively. As stated in the paper, NDS (Row 5) has a comparable performance comparing to the NSS strategy (Row 6), yet the performance is greatly reduced if many inclined walls appeared in a floor plan (Column 3) since most of the inclined walls are suppressed. }\label{Fig:example_outputs}
\end{figure*}

\begin{figure*}[!t]
    \centering
    \includegraphics[width=.75\linewidth]{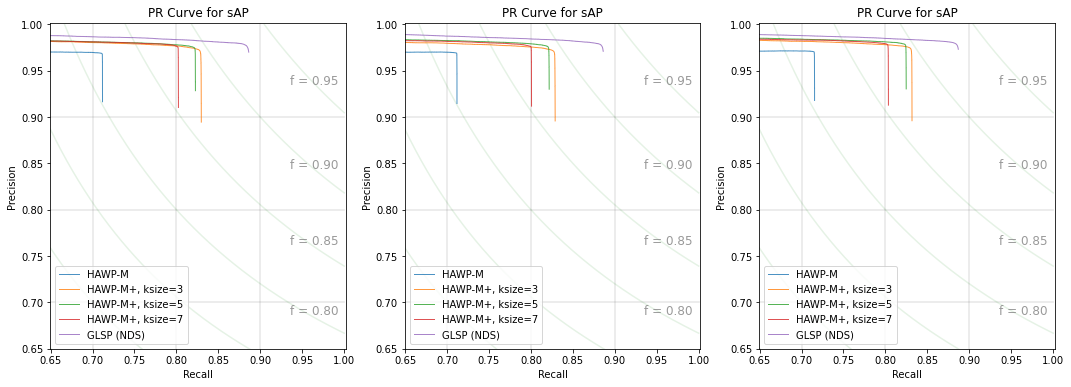}
    \caption{Precision-Recall (PR) curves of all three junction detection experiments.}\label{Fig:sup_junction}
\end{figure*}

\begin{figure*}[!t]
    \centering
    \begin{subfigure}[c]{.99\linewidth}
        \centering
        \includegraphics[width=.99\linewidth]{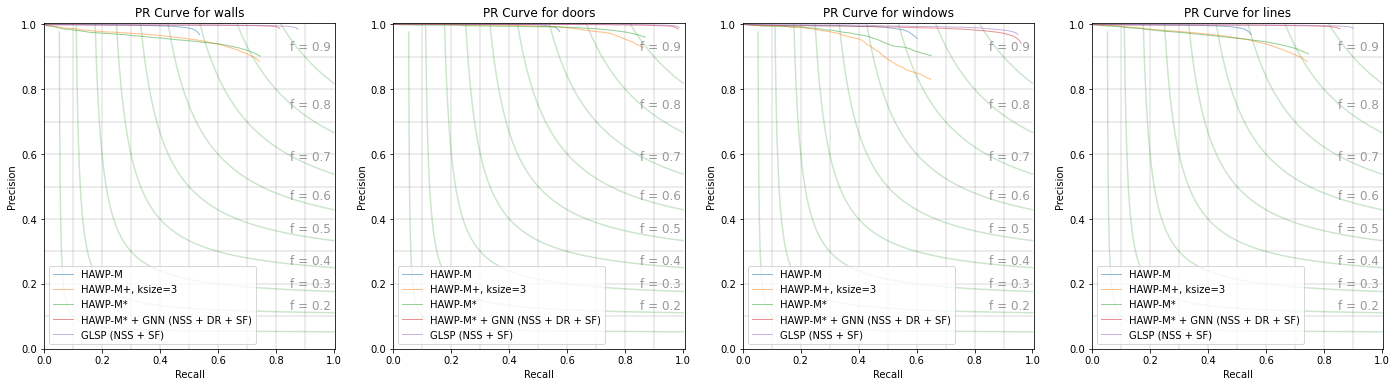}
        \caption{}\label{Fig:sup_overview_pr_line_1}
    \end{subfigure}
    \hfill
    \begin{subfigure}[c]{.99\linewidth}
        \centering
        \includegraphics[width=.99\linewidth]{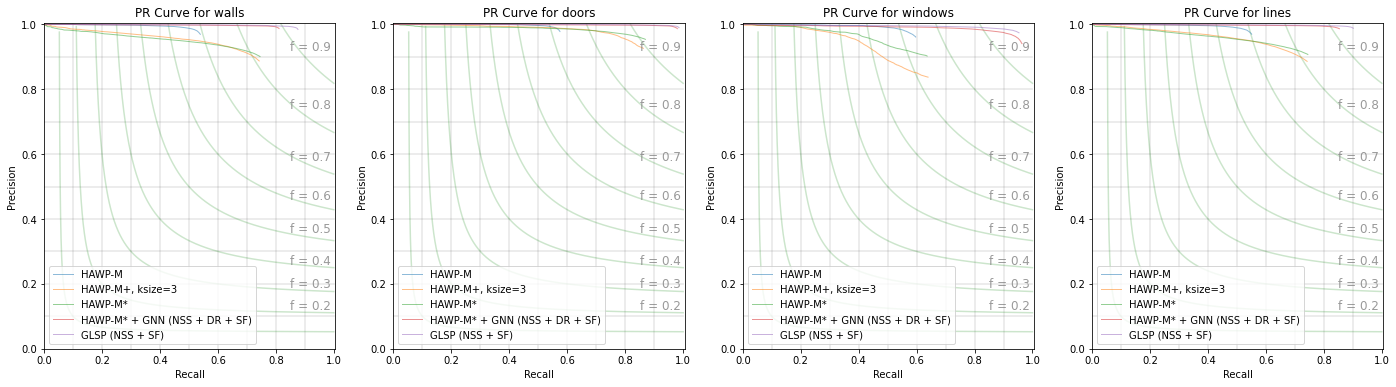}
        \caption{}\label{Fig:sup_overview_pr_line_2}
    \end{subfigure}
    \hfill
    \begin{subfigure}[c]{.99\linewidth}
        \centering
        \includegraphics[width=.99\linewidth]{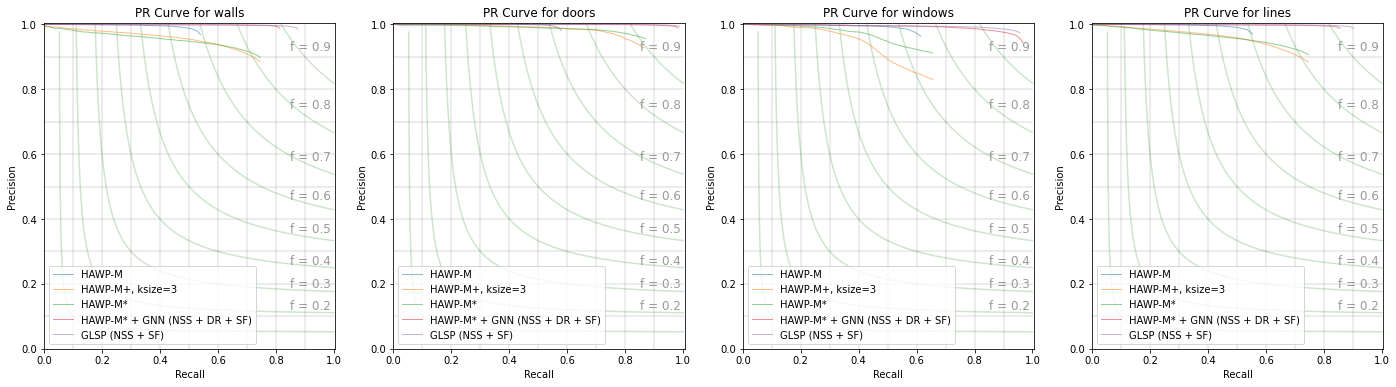}
        \caption{}\label{Fig:sup_overview_pr_line_3}
    \end{subfigure}
    \caption{Precision-Recall (PR) curves of all three line segment detection experiments.}\label{Fig:sup_overview}
\end{figure*}

\begin{figure*}[!t]
    \centering
    \begin{subfigure}[c]{.99\linewidth}
        \centering
        \includegraphics[width=.99\linewidth]{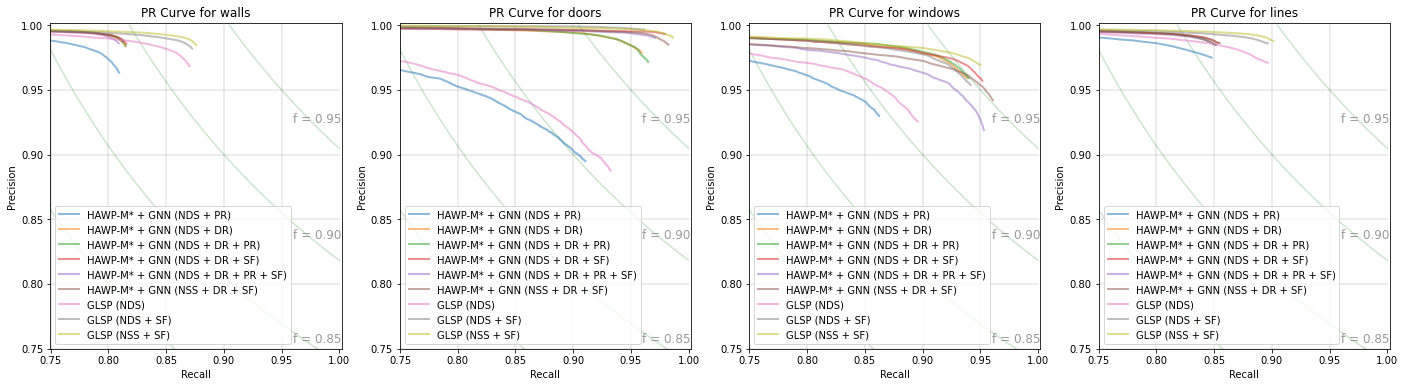}
        \caption{}\label{Fig:sup_line_pr_line_1}
    \end{subfigure}
    \hfill
    \begin{subfigure}[c]{.99\linewidth}
        \centering
        \includegraphics[width=.99\linewidth]{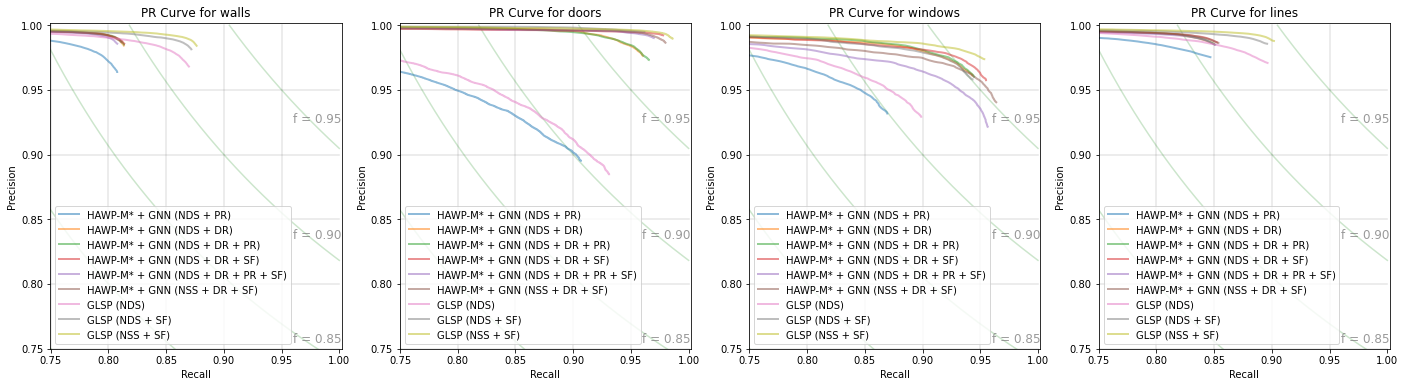}
        \caption{}\label{Fig:sup_line_pr_line_2}
    \end{subfigure}
    \hfill
    \begin{subfigure}[c]{.99\linewidth}
        \centering
        \includegraphics[width=.99\linewidth]{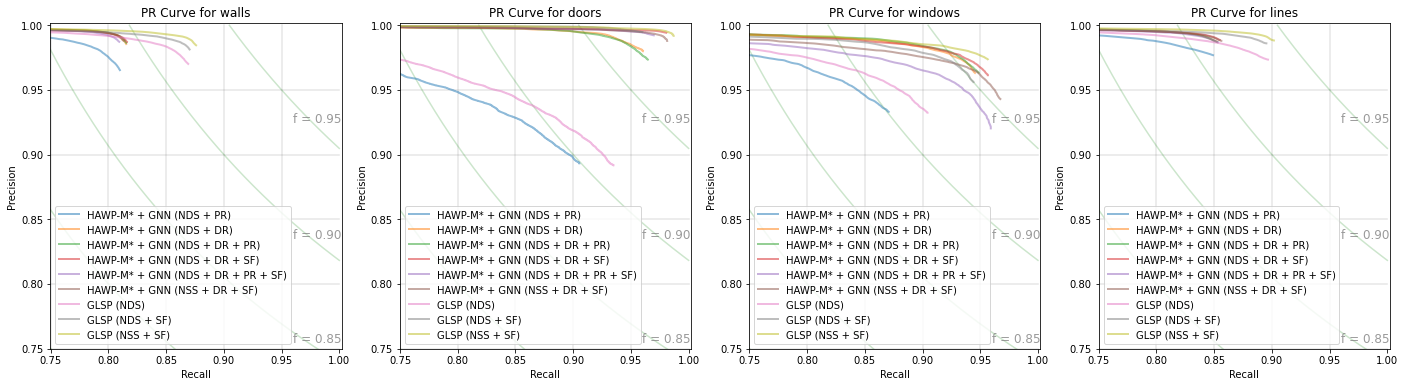}
        \caption{}\label{Fig:sup_line_pr_line_3}
    \end{subfigure}
    \hfill
    \begin{subfigure}[c]{.99\linewidth}
        \centering
        \includegraphics[width=.99\linewidth]{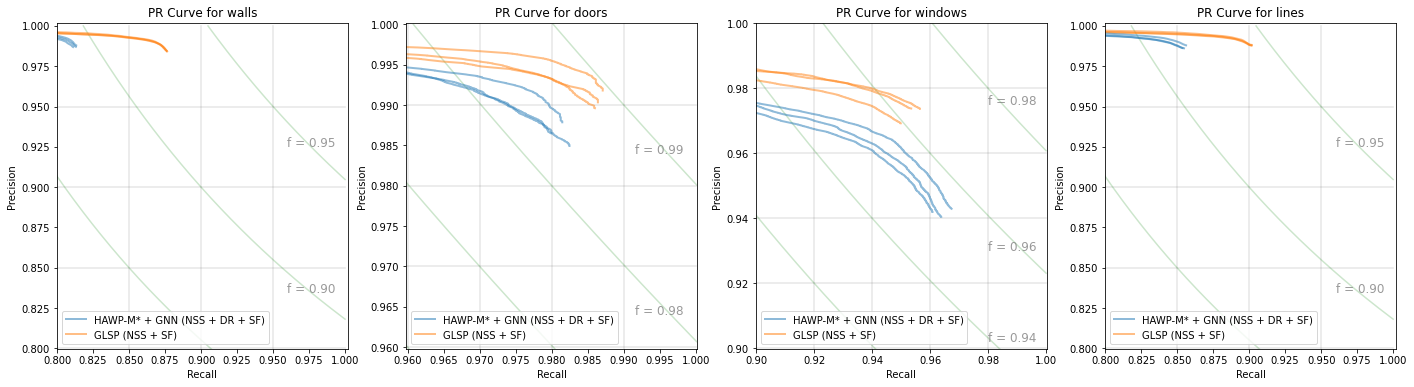}
        \caption{}\label{Fig:sup_best_line_pr_line}
    \end{subfigure}
    \caption{(a)-(c) Precision-Recall (PR) curves of all three experiments for GLSP as integratable modules and end-to-end GLSPs. (d) PR curves of all three experiments for the best integratable module and the best end-to-end GLSP. To clarify the difference between curves, the scales of the figures in (d) are not identical.}\label{Fig:sup_line_pr_line}
\end{figure*}

\begin{figure*}[!t]
    \centering
    \begin{subfigure}[c]{.99\linewidth}
        \centering
        \includegraphics[width=.99\linewidth]{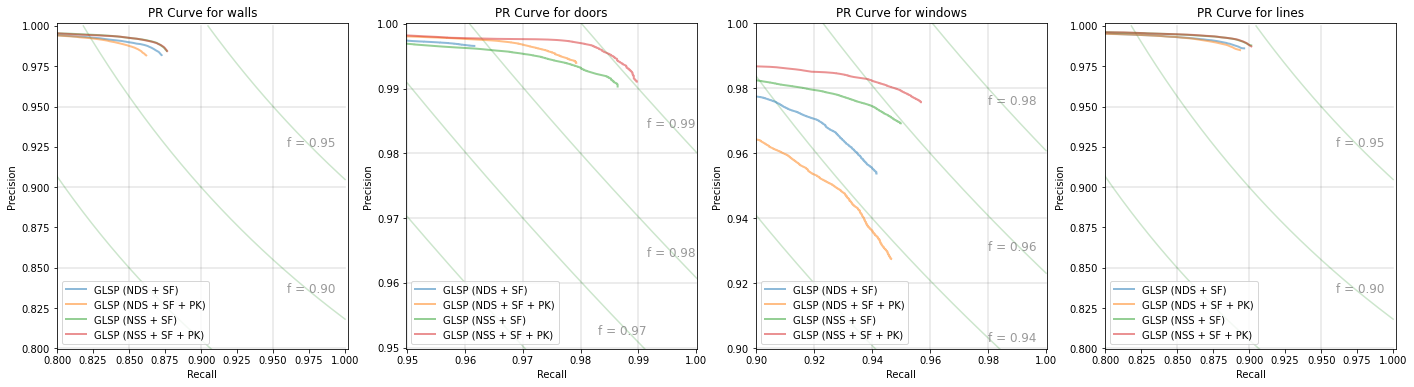}
        \caption{}\label{Fig:sup_prior_pr_line_1}
    \end{subfigure}
    \hfill
    \begin{subfigure}[c]{.99\linewidth}
        \centering
        \includegraphics[width=.99\linewidth]{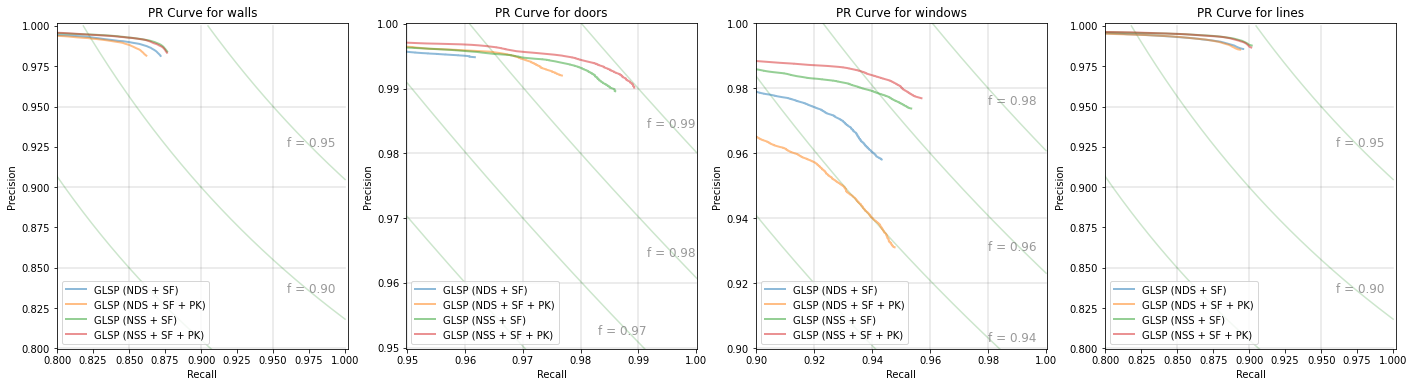}
        \caption{}\label{Fig:sup_prior_pr_line_2}
    \end{subfigure}
    \hfill
    \begin{subfigure}[c]{.99\linewidth}
        \centering
        \includegraphics[width=.99\linewidth]{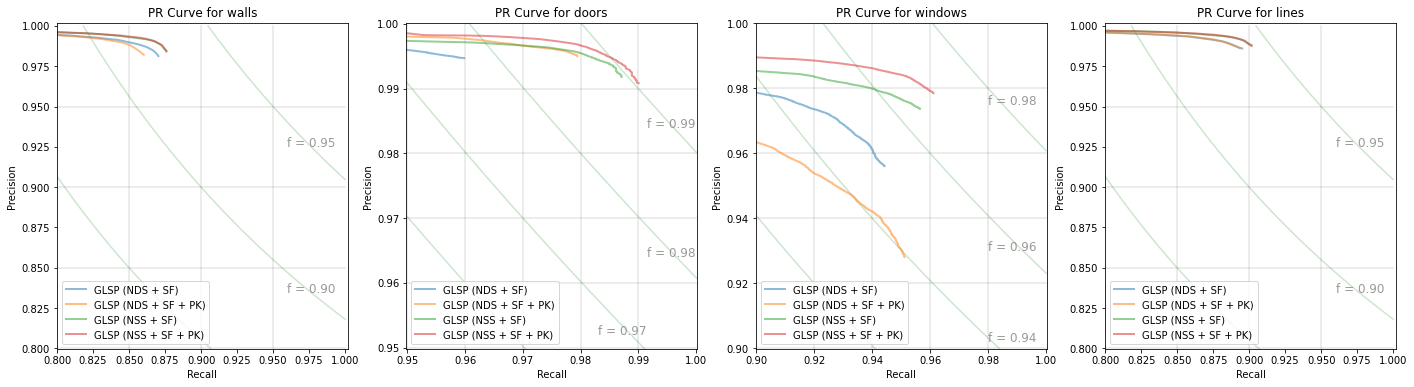}
        \caption{}\label{Fig:sup_prior_pr_line_3}
    \end{subfigure}
    \caption{Precision-Recall (PR) curves of all three line segment detection experiments with prior knowledge. To clarify the difference between curves, the scales of the figures are not identical.}\label{Fig:sup_prior}
\end{figure*}

We show line segment detection qualitative results from GLSPs and variants of HAWP in Figure \ref{Fig:example_outputs}. Images appeared on the top row are the inputs. The results of sAP for walls, doors, windows, are shown in Table \ref{tab:sup_line_seg_eval_wall}, \ref{tab:sup_line_seg_eval_door}, and \ref{tab:sup_line_seg_eval_win}, respectively. Figure \ref{Fig:sup_junction}-\ref{Fig:sup_prior} are the PR curves of all three experiments for junction detection, line segment detection, ablation studies, and the influence of introducing prior knowledge, respectively.

\begin{table}[!htp]
    \centering
    \caption{Quantitative evaluation of detecting line segments of walls.}
    \setlength{\tabcolsep}{1em}
    \resizebox{0.99\linewidth}{!}{%
    \begin{tabular}{l c c c }
    \toprule
        & sAP$^{8}_{wall}$ & sAP$^{16}_{wall}$ & sAP$^{32}_{wall}$ \\
        \midrule
        HAWP-M & $52.97 \textcolor{blue}{\pm 0.17}$ & $53.65 \textcolor{blue}{\pm 0.13}$ & $53.87 \textcolor{blue}{\pm 0.10}$ \\
        \cmidrule(lr){2-4}
        HAWP-M+ & $69.29 \textcolor{blue}{\pm 0.07}$ & $71.36 \textcolor{blue}{\pm 0.05}$ & $72.18 \textcolor{blue}{\pm 0.06}$ \\
        \cmidrule(lr){2-4}
        HAWP-M* & $69.11 \textcolor{blue}{\pm 0.07}$ & $71.33 \textcolor{blue}{\pm 0.08}$ & $72.25 \textcolor{blue}{\pm 0.09}$ \\
        \cmidrule(lr){2-4}
        HAWP-M* + GNN & $\underline{78.13} \textcolor{blue}{\pm 0.21}$ & $\underline{81.04} \textcolor{blue}{\pm 0.11}$ & $\underline{81.43} \textcolor{blue}{\pm 0.08}$ \\
        \cmidrule(lr){2-4}
        GLSP & $\textbf{84.88} \textcolor{blue}{\pm 0.17}$ & $\textbf{87.42} \textcolor{blue}{\pm 0.04}$ & $\textbf{87.79} \textcolor{blue}{\pm 0.03}$ \\
    \toprule
    \end{tabular}
    }
    \label{tab:sup_line_seg_eval_wall}
\end{table}

\begin{table}[!htp]
    \centering
    \caption{Quantitative evaluation of detecting line segments of doors.}
    \setlength{\tabcolsep}{1em}
    \resizebox{0.99\linewidth}{!}{%
    \begin{tabular}{l c c c }
    \toprule
        & sAP$^{8}_{door}$ & sAP$^{16}_{door}$ & sAP$^{32}_{door}$ \\
        \midrule
        HAWP-M & $56.66 \textcolor{blue}{\pm 0.29}$ & $57.49 \textcolor{blue}{\pm 0.22}$ & $57.70 \textcolor{blue}{\pm 0.22}$ \\
        \cmidrule(lr){2-4}
        HAWP-M+ & $84.38 \textcolor{blue}{\pm 0.24}$ & $85.32 \textcolor{blue}{\pm 0.24}$ & $85.71 \textcolor{blue}{\pm 0.28}$ \\
        \cmidrule(lr){2-4}
        HAWP-M* & $84.72 \textcolor{blue}{\pm 0.18}$ & $85.94 \textcolor{blue}{\pm 0.14}$ & $86.45 \textcolor{blue}{\pm 0.18}$ \\
        \cmidrule(lr){2-4}
        HAWP-M* + GNN & $\underline{94.47} \textcolor{blue}{\pm 0.18}$ & $\underline{98.01} \textcolor{blue}{\pm 0.12}$ & $\underline{98.52} \textcolor{blue}{\pm 0.11}$ \\
        \cmidrule(lr){2-4}
        GLSP & $\textbf{97.50} \textcolor{blue}{\pm 0.19}$ & $\textbf{98.59} \textcolor{blue}{\pm 0.06}$ & $\textbf{98.75} \textcolor{blue}{\pm 0.03}$ \\
    \toprule
    \end{tabular}}
    \label{tab:sup_line_seg_eval_door}
\end{table}

\begin{table}[!htp]
    \centering
    \caption{Quantitative evaluation of detecting line segments of windows.}
    \setlength{\tabcolsep}{1em}
    \resizebox{0.99\linewidth}{!}{%
    \begin{tabular}{l c c c }
    \toprule
        & sAP$^{8}_{win}$ & sAP$^{16}_{win}$ & sAP$^{32}_{win}$ \\
        \midrule
        HAWP-M & $56.73 \textcolor{blue}{\pm 0.75}$ & $60.06 \textcolor{blue}{\pm 0.75}$ & $60.88 \textcolor{blue}{\pm 0.77}$ \\
        \cmidrule(lr){2-4}
        HAWP-M+ & $58.49 \textcolor{blue}{\pm 0.68}$ & $61.38 \textcolor{blue}{\pm 0.66}$ & $62.63 \textcolor{blue}{\pm 0.60}$ \\
        \cmidrule(lr){2-4}
        HAWP-M* & $59.55 \textcolor{blue}{\pm 0.70}$ & $62.71 \textcolor{blue}{\pm 0.86}$ & $64.21 \textcolor{blue}{\pm 0.65}$ \\
        \cmidrule(lr){2-4}
        HAWP-M* + GNN & $\underline{84.97} \textcolor{blue}{\pm 0.74}$ & $\textbf{95.45} \textcolor{blue}{\pm 0.37}$ & $\textbf{97.55} \textcolor{blue}{\pm 0.08}$ \\
        \cmidrule(lr){2-4}
        GLSP & $\textbf{88.44} \textcolor{blue}{\pm 0.81}$ & $\underline{94.77} \textcolor{blue}{\pm 0.33}$ & $\underline{96.16} \textcolor{blue}{\pm 0.13}$ \\
    \toprule
    \end{tabular}}
    \label{tab:sup_line_seg_eval_win}
\end{table}

\end{appendices}

\end{document}